\documentclass[12pt,number,preprint,times,letterpaper]{elsarticle}

\usepackage{algorithm,algorithmic}

\usepackage{amsfonts,amsmath,amssymb,amsthm,mathrsfs,mathtools,dsfont}
\usepackage{subfig}

\usepackage{breakcites}

\usepackage{booktabs}

\usepackage[
  breaklinks=true,
]{hyperref}
\hypersetup{
  pdfauthor={\ldots},
  % ...
}
\usepackage{bookmark}
\hypersetup{bookmarksopen=true}

\usepackage{listings}
\usepackage[usenames,dvipsnames]{color}

\biboptions{authoryear}

\journal{ }

\usepackage{accents}

\usepackage{enumitem}

\usepackage{pifont}
\newcommand{\interior}[1]{%
  {\kern0pt#1}^{\mathrm{o}}%
}

\usepackage{fancyhdr}
\begin{document}

\begin{frontmatter}
  \title{Predicting Ethnicity from Names with \texttt{rethnicity}: Methodology and Application}

  \author{Fangzhou Xie\fnref{l1}} 
  \ead{fangzhou.xie@rutgers.edu}

  \fntext[l1]{New Jersey Hall, Room 202, 75 Hamilton Street, New Brunswick, NJ 08901.}
  \address{Department of Economics, Rutgers University}

  \date{\today}

  \begin{abstract}
    In this study, a new R package, \texttt{rethnicity}\footnote{
      \href{https://github.com/fangzhou-xie/rethnicity}{https://github.com/fangzhou-xie/rethnicity}.
      It has also been published on \href{https://cran.r-project.org/web/packages/rethnicity/index.html}{CRAN}.
    } is provided for predicting ethnicity based on names.
    The Bidirectional LSTM and Florida Voter Registration were used as the model and training data, respectively. Special care was given for the accuracy of minority groups,
    by adjusting the imbalance in the dataset.
    The models were trained and exported to C++ and then integrated with R using Rcpp.
    Additionally, the availability, accuracy, and performance of the package were compared with other solutions.
  \end{abstract}



\end{frontmatter}


\section{Introduction}

The study on the differential effects of ethnicity requires researchers to have
ethnic information available in a dataset.
However, such information is usually not readily available\footnote{
  Health care is one of the areas that must be studied for ethnic disparities
  in insurance plans, and researchers in this field should deal with the missing
  ethnic information~\citep{fiscella2006}.
}.
When only names are available in the dataset, one naturally wants to predict
people's ethnicity based on their names, as names are usually highly correlated with
their races.

In fact, surname analysis has been used for many years to identify ethnicity\footnote{
  See~\cite{fiscella2006} for a survey.
}, but the application of deep learning can make it even simpler,
as illustrated by~\cite{sood2018}.

Herein, a novel approach to predict ethnicity from names is proposed and
an R package is provided \texttt{rethnicity}\footnote{
  https://github.com/fangzhou-xie/rethnicity
}.
The developed method achieves a good performance, is fast, and free.

The rest of the article is organized as follows.
Section~\ref{sec:methodology} describes the methodology of the package.
Section~\ref{sec:data} discusses some of the model’s implementation details.
Section~\ref{sec:feature} highlights the notable features unique to the package.
Section~\ref{sec:comparison} compares its availability, accuracy, and performance
against other solutions.
Section~\ref{sec:application} presents an example as a code snippet that can be used to analyze racial differences in political donations.
Finally, Section~\ref{sec:conclusion} concludes the study.


\section{Methodology}\label{sec:methodology}

This section introduces the methodology of the prediction method
provided by the R package and the procedures used to develop it.

\subsection{Undersampling for the Imbalanced Racial Distribution}\label{sec:methodology-undersampling}

Most classification algorithms assume a relatively balanced dataset and
equal misclassification cost~\citep{sun2009}. When applying
them on imbalanced data, where the instances of some classes are
significantly larger or smaller than other classes, the algorithm will
mainly focus on the majority class and hence ignore the minority
classes. One example is fraud detection, where most of the transactions
are normal, but a few are fraudulent~\citep{fawcett1997}. 

This is also a concern in our application. An attempt is made
to predict ethnic groups from people's names, which is a natural example of
classifying imbalanced data\footnote{
  The dataset from Florida Voter Registration~\citep{sood2017} was used in this study,
  and the predicted ethnic groups are defined in the U.S. context.
  The same methodology can be applied to build a classifier
  for another country or region if a proper dataset
  with both names and races is available.
}.

To overcome this problem, one
important method is to oversample the minority class
\citep{chawla2002,fernandez2018}. However, in this study, because
abundant data points are available, the
majority classes were undersampled to achieve a balanced dataset. This will also help
reduce the training and testing time for the model owing to a large
dataset\footnote{
  Details in Section~\ref{sec:data-source}.
}.

Furthermore, first names are not only associated with gender,
but also with race~\citep{fryer2004}.
Hence, the dataset was grouped by both ethnicity and gender,
and all the groups except for the smallest one were undersampled.
Two different models were trained for classifying
ethnicity. One was trained using only last names, while the other was trained on first names.
Therefore, it is crucial to adjust the dataset based on both gender and race
to avoid disproportionate classification errors on minority classes.

\subsection{Character-level Dictionary}\label{sec:methodology-dictionary}

Classic natural language processing (NLP) models consider ``tokens'' to
be the building blocks of languages. This appears natural to humans
because words and phrases are considered the smallest elements of language in
our daily use. However, for algorithms to process sentences, there is a need to tokenize them,
create a vocabulary, and then build a model based on the vocabulary. This process will become cumbersome as the size of the data increases and there is a need to
retain an extremely large vocabulary, where some tokens are very
common, while several are extremely infrequent~\citep{zipf1936}.

Moreover, the vocabulary is usually built during model training,
and out-of-vocabulary (OOV) tokens may exist during inference once the model
is deployed. The usual practice is to map OOV tokens to an ``unknown'' token
and consider tokens that are not seen in the training process to be similar.
This will inevitably lead to information loss, as some tokens
are extremely informative but also infrequent
(e.g., \ special words or abbreviations with domain knowledge).

To overcome this, efforts have been taken in the machine learning field to build
models directly on characters, instead of
tokens~\citep{zhang2015,sutskever2011}\footnote{These
  character-level models only consider 26 English letters (and some
  symbols). Recently,~\cite{xue2021} proposed a byte-based model, which is
  fully compliant with the UTF-8 standard and can deal with
  non-English texts.
}.
It is easier to enumerate all possible characters\footnote{
  In the case of English, only 26 letters are needed in addition to symbols when necessary.
  A larger dictionary could be used by including upper-case letters.
  It is more efficient to use lower-case letters for classifying names,  as upper-case letters will be fewer and the model may not have
  enough opportunities to learn from the upper-case letters.
}
and maintain a dictionary
of those characters, rather than a dictionary consisting of distinct tokens.
Hence, in this study, OOV problems are not a concern because all the characters have been kept in
the dictionary.

The other benefit of using a
character-level dictionary is that it significantly reduces the size of
the dictionary and the parameters needed in the
model, as many, if not most, parameters in are needed in NLP for
capturing the meaning of tokens~\citep{xue2021}.
In this way, a model can be made lightweight without sacrificing accuracy
and also gain some speed in the inference stage.

\subsection{Bidirectional LSTM}\label{sec:methodology-bilstm}

Long short-term memory~\citep[LSTM]{hochreiter1997} has been widely used in sequence modeling since its proposal\footnote{
  Some of the most recent and exciting developments in LSTM include BERT~\citep{devlin2019} and its variants.
}.
Moreover,~\cite{graves2005} proposed bidirectional LSTM (BiLSTM), which captures the context
even better than the unidirectional LSTM model\footnote{
  BiLSTM uses both forward-and backward-passing LSTMs to capture the context
  of sequential data, which is one of the reasons it works well.
  Although BiLSTM cannot be used for real-time prediction, this is less of a concern
  for our name classification task.
}.

In this package, BiLSTM is used as the model architecture for predicting race from names.
The model was built with 256 units of an embedding layer and four BiLSTM layers with 512 units each.
The final output layer is a dense layer of four units (equal to the number of races for the classification problem)
with softmax activation function\footnote{
  The test accuracy is given in~\ref{sec:comparison-accuracy}.
}\textsuperscript{,}\footnote{
  Last name and full name models have the same architecture but differ in terms of the dataset used for training.
}.

The performance of the model in terms of accuracy is listed in Table~\ref{tab:teacher-model}.

\subsection{Distillation of Knowledge}\label{sec:methodology-distillation}

However, training models with a character-level dictionary might be
more difficult than token-level dictionary models.
This might be the only drawback of using a character-level dictionary.
To overcome this difficulty,
a large BiLSTM model with many parameters is trained for better accuracy.
However, this trained model is extremely large and would be difficult to deploy
in production. Therefore, ``model distillation'' is used to compress the information into a smaller model.

To compress a model,~\cite{hinton2015} proposed the ``distillation'' technique for extracting information
from large models and teaching a smaller model to achieve a similar prediction\footnote{
  or for using the prediction of an ensemble of models to distill information into a single
  small model.
}. To be more precise, the ``student'' model is trained to match the ``teacher'' model, and the knowledge
is transferred from the teacher to the student.
In this way, the student will ``learn'' the interclass relationship better than directly learning
from the data.

The distillation trick is applied on the trained large model to obtain a smaller model
with the same architecture but fewer parameters and layers\footnote{
  The architecture of the student model includes dense, BiLSTM, and dense layers.
  However, there are only 32 units for dense, 64 units each for the two layers of BiLSTM\@.
}. The smaller model is compressed from the larger model and becomes the model used for inference in production\footnote{
  The accuracy comparison is shown in Section~\ref{sec:comparison-accuracy}.
}.

Table~\ref{tab:student-model} shows the test accuracy of the student (smaller) model.

\subsection{Export to C++}

After training the teacher and student models, the student model is exported to C++
via \texttt{frugally-deep}\footnote{
  \url{https://github.com/Dobiasd/frugally-deep}
} project.
Hence, the model is no longer dependent on the installation
of \texttt{Keras} (or \texttt{tensorflow}). Subsequently, the model is loaded directly
in C++ with very few dependencies\footnote{
  The \texttt{frugally-deep} is a lightweight header-only C++ project that depends only on
  FunctionalPlus, Eigen, and Json projects, which are all header-only projects.
}.

To make the model callable from R, an interface must be developed using Rcpp~\citep{eddelbuettel2011}.
This will provide a wrapper around the underlying C++ code for loading the model
and run the prediction for the names.
Additionally, the prediction can be parallelly done by multi-threading.
These features will enable the names to be processed rapidly for the prediction
of ethnicity.

\section{Data and Preprocessing}\label{sec:data}

\subsection{Name and Ethnicity Data}\label{sec:data-source}

To train the ethnicity classification model using names, we need a dataset that includes names
as well as individual-level racial information.
Fortunately, the Florida Voter Registration Dataset~\citep{sood2017}
is an excellent candidate for this purpose.

However, the dataset contains names and races for almost 13 million people in Florida,
and the racial distribution is naturally imbalanced.
First, the names of Native Americans\footnote{
  These names are defined in the Florida Voter Registration dataset as
  ``American Indian or Alaskan Native.’’
} and multi-racial names were dropped because these groups have little data.
Furthermore,
Asian or Pacific Islanders, Hispanic, Non-Hispanic Black, and Non-Hispanic White
were defined as the four categories for the classification problem\footnote{
  Henceforth, these groups will be referred to as Asian, Hispanic, Black, and White.
}.

Table~\ref{tab:name-freq} lists the frequency of names
grouped by ethnicity and gender.
The undersampling procedure discussed
in Section~\ref{sec:methodology-undersampling} takes the smallest group, namely, Asian Male,
and randomly selects samples for all other groups to eventually have the same group size.
After undersampling, each race-gender group contains
the same number (i.e., 104,632) of names.

\begin{table}[ht]
  \centering
  \begin{tabular}{ccrr}
    \toprule
    Race     & Gender & Count Before & Count After \\
    \midrule
    Asian    & Female & 131602       & 104632      \\
    Asian    & Male   & 104632       & 104632      \\
    Black    & Female & 989142       & 104632      \\
    Black    & Male   & 717118       & 104632      \\
    Hispanic & Female & 1137594      & 104632      \\
    Hispanic & Male   & 925623       & 104632      \\
    White    & Female & 4419030      & 104632      \\
    White    & Male   & 3963833      & 104632      \\
    \bottomrule
  \end{tabular}
  \caption{Count of Names, Grouped by Ethnicity and Gender.}\label{tab:name-freq}%
\end{table}

\subsection{Character Encoding}\label{sec:data-encoding}

For the characters to be processed by the algorithm,
characters must be encoded using numeric values. Because a character-level dictionary is the focus of this study,
\footnote{
  The 26 lower-case English letters, a space character (`` ''),
  an empty character (``E''), and an unknown character (``U'') are considered.
  Hence, the size of the dictionary is 29
},
the dictionary is small and pre-defined.
It is possible to map the characters from all names to a value in the dictionary.

In practice, first, upper-case letters are mapped to their lower-case counterparts,
then all the punctuation symbols to space,
and, finally, all other characters to ``U'' (Unknown). Furthermore, to ensure that all input data are of
equal length, ``E'' (Empty) is inserted for names shorter than the threshold,
and the additional characters in the names longer than the threshold are trimmed\footnote{
  The threshold is ten for each name component.
  In other words, for the last name model, the input length is ten.
  The full name model takes both the first and last names, each being ten, and
  the final length is thus 20.
}. Subsequently, the input name is transformed into a vector of integers according to the
mapping described in Section~\ref{sec:methodology-dictionary}.

\subsection{Sequence Padding and Alignment}\label{sec:data-alignment}

The number of characters in people’s names varies, as does their encoded numeric
representation. However, the model will only accept input vectors of equal length.
To achieve this, it is necessary to determine the expected length of a fixed number of
input vectors and process the input data such that each string has the same length.
Longer strings will be truncated and shorter ones will be padded\footnote{
  This process is called ``padding'' and is the usual practice in the preprocessing of RNN models.
}.

Therefore, after padding the sequences at 10, all surnames will have the same length after the padding process.
For example, consider the last name ``Smith.'' First, the last name is converted to lower case to obtain
``smith,'' and then five Empty (``E'') characters are added to get a vector of length ten. By contrast, for ``Christensen,''
the name is first converted into lower case, and then the last ``n'' is trimmed.

For the model that leverages both the first and last names, special care must be taken.
Because first names also vary in length, if both the first and last names is considered as a single string,
the starting point of the last name will also vary\footnote{
  This fact makes the full-name model insensitive to the names with first and last names from
  different ethnic groups, for example, the descendants of immigrants. Early models trained without adjusting the position
  of the last names tend to focus more on the first names instead of the last names and tend to predict
  ``Andrew Yang'' as White instead of Asian. Hence, there is a need to introduce an alignment procedure.
}.
The solution would be to pad the first and last names separately and then concatenate them into a single
vector. This guarantees the same starting position for all last names across the sample.
The first and last names are both padded with ten characters so that the total concatenated input
for the full name model is 20, but the last name always starts at position 11.

The alignment method allows the model to recognize the difference between the first and last names,
and the accuracy is higher compared with the models trained without the alignment process.

\section{Features of the Package}\label{sec:feature}

In this section, several notable features of the package are presented.

\subsection{Native in R}\label{sec:feature-native-in-r}

The R community has incorporated mature deep learning libraries from
Python~\citep{falbel2021,kalinowski2021} via
\texttt{reticulate} package~\citep{ushey2021}.
In essence, users need
to install a separate Python environment with the required libraries
(e.g.,~tensorflow), and then \texttt{reticulate} would provide the
interface from R to Python so that neural networks can be developed in R.

In practice, this could be problematic for researchers, as this approach
heavily relies on another language (Python) and may cause issues while replicating
the studies. Moreover, in most applied research, researchers only need to make
inferences on the dataset at hand. Installing and maintaining these
libraries\footnote{
  The installation processes of these libraries and
  their dependencies are not trivial. Even after correct installation,
  they take up multiple gigabytes of storage.
} are usually problematic, even for veterans.

To ease the burden on users, the modeling processes and
deployment are separated so that end-users would only need to install minimal
dependencies on their machines.
After training the teacher and distilling the student model,
the student model is exported to C++ via \texttt{frugally-deep}.
At this stage, the model is no longer dependent on any of the deep
learning frameworks. Furthermore, to run the inference of this model in R,
Rcpp is used to provide an interface to the C++ model and make it callable from
R. The package is henceforth considered to be lightweight, without the need
to refer to external languages (e.g.,~Python) by those who need to predict
ethnicity from names.

\subsection{Mature Dependencies}\label{sec:feature-dependency}

The \texttt{torch} package~\citep{falbel2021a} aims to build the
\texttt{PyTorch} package native to R, which is in sharp contrast
to the approach of \texttt{tensorflow} and \texttt{keras} for R.
However, this project is experimental and not ready for production.
Additionally, this approach
requires the installation of the massive \texttt{torch}
package. Furthemore, such an installation might not be required for many people.

For the \texttt{rethnicity} package, there are only three dependencies:
\texttt{Rcpp}~\citep{eddelbuettel2011},
\texttt{RcppThread}~\citep{nagler2021},
and \texttt{RcppEigen}~\citep{bates2013}. They are
all being well-tested, with mature packages published on CRAN and widely
used by many other packages in the R community. In particular, \texttt{Rcpp}
provides an interface to C++, \texttt{RcppThread} provides the
multi-threading support for fast inference, and \texttt{RcppEigen}
provides a fast and efficient matrix computation of the neural network\footnote{
  Apart from these three packages, it also depends on \texttt{frugally-deep}
  and its dependencies. But these header-only dependencies are
  included in the package during installation, and there is no need
  to link against external packages.
}.

\subsection{High Performance}\label{sec:feature-performance}

With the rise of empirical economics, economists often find themselves
dealing with much larger datasets~\citep{einav2014}.
Therefore, it is critical to have packages designed considering the performance requirements to process datasets as quickly as possible.

In terms of predicting ethnicity/nationality from names,
some good API services\footnote{For example, \texttt{nationalize.io}
  (\url{https://nationalize.io/}) and \texttt{NamePrism}
  (\url{https://www.name-prism.com/}).} exist. However, as is the case with
most API services, they are rate-limited to make their service reliable
and sustainable. This may create bottlenecks for researchers who have a
large collection of names waiting to be predicted.

The \texttt{rethnicity} package is built with this concern in mind.
By leveraging C++ and multi-threading\footnote{
  This is possible because the models exported by \texttt{frugally-deep} are thread-safe.
},
the package could achieve extremely fast inference speeds.
To illustrate this point, the inference speed between
\texttt{rethnicity} and \texttt{ethnicolr} is compared in Section~\ref{sec:comparison-performance}.

\section{Comparison with Existing Packages}\label{sec:comparison}

\subsection{Availability}\label{sec:comparison-availability}

Table~\ref{tab:comparison} shows the differences between \texttt{rethnicity} and other solutions
for predicting ethnicity from names.
The comparison is made on four aspects: cost, rate limit, dependencies,
and language.

\begin{table}[ht]
  \centering
  \begin{tabular}{ccccc}
    \toprule
               & Ethnicity & Ethnicolr & NamePrism & nationalize.io \\
    \midrule
    Cost       & free      & free      & free      & paid           \\
    Rate Limit & No        & No        & Yes       & Yes            \\
    Dependency & Low       & High      & N/A       & N/A            \\
    Language   & R         & Python    & API       & API            \\
    \bottomrule
  \end{tabular}
  \caption{
    Comparison across some publicly available services/packages for predicting ethnicity from names.
    \texttt{rethnicity} provides a free and light-weight package for the R community
    without rate-limiting.
  }\label{tab:comparison}
\end{table}

\texttt{NamePrism} is free but is rate-limited to 60 requests per minute.
\texttt{nationalize.io} offers 1000 free requests each day and requires a subscription to their services for more
names to be processed in a day.
\texttt{ethnicolr} might be the most similar in scope as \texttt{rethnicity},
However, it is written in Python and requires the installation of TensorFlow to run the inference.
The installation of tensorflow might be daunting for many who only want to run the inference
on a name dataset they have.

In general, this study aimed to develop a simple, easy, and free package for use in the R community
with guaranteed accuracy and performance, which is achieved by the \texttt{rethnicity} package.

\subsection{Accuracy}\label{sec:comparison-accuracy}

Tables~\ref{tab:teacher-model} and~\ref{tab:student-model} present the prediction accuracy on the
test data not included during the training process. Table~\ref{tab:teacher-model} shows the accuracy of the trained teacher model.
, and Table~\ref{tab:student-model} shows the accuracy of the student model.
Note that the full name model performs better than the one that only leverages last name information,
for both teacher and student models. Additionally, the precision of the student model degrades compared to the
teacher model but is still sufficiently high and close to that of the teacher.

Moreover, if we compare the results within each ethnic group, the accuracy for each group
is roughly balanced, and the performance is slightly better for the minority groups.
This is the case for both teacher and student models.
This suggests that the undersampling approach used in this study to adjust the imbalance
(described in Section~\ref{sec:methodology-undersampling}) in the dataset works well.
If the results are compared to \texttt{ethnicolr}~\citep{sood2018},
\texttt{rethnicity} shows significantly better results on the prediction of Asian, Hispanic, and black people,
albeit the precision is lower for white people\footnote{
  The accuracies in~\cite{sood2018} are disproportionately high for white people,
  which might suggest that the classifier tends to always predict white to minimize loss.
  Therefore, adjustments are performed for an imbalanced dataset.
}.

\begin{table}[ht]
  \begin{tabular}{rrrrrrrr}
    \toprule
             & \multicolumn{3}{c}{\bfseries Full name } & \multicolumn{3}{c}{\bfseries Lastname} &                                                    \\
    \cmidrule(lr){2-4}\cmidrule(lr){5-7}\cmidrule(lr){8-8}
             & precision                                & recall                                 & f1-score & precision & recall & f1-score & support \\
    \cmidrule(lr){2-4}\cmidrule(lr){5-7}\cmidrule(lr){8-8}
    asian    & 0.87                                     & 0.76                                   & 0.81     & 0.87      & 0.69   & 0.77     & 41861   \\
    black    & 0.74                                     & 0.77                                   & 0.76     & 0.65      & 0.80   & 0.72     & 41904   \\
    hispanic & 0.86                                     & 0.87                                   & 0.86     & 0.84      & 0.85   & 0.85     & 41940   \\
    white    & 0.67                                     & 0.73                                   & 0.70     & 0.62      & 0.58   & 0.60     & 41707   \\
    total    & 0.79                                     & 0.78                                   & 0.78     & 0.74      & 0.73   & 0.73     & 167412  \\
    \bottomrule
  \end{tabular}
  \caption{Accuracy on the test data for the teacher model before distillation.}\label{tab:teacher-model}
\end{table}

\begin{table}[ht]
  \begin{tabular}{rrrrrrrr}
    \toprule
             & \multicolumn{3}{c}{\bfseries Full name } & \multicolumn{3}{c}{\bfseries Lastname} &                                                    \\
    \cmidrule(lr){2-4}\cmidrule(lr){5-7}\cmidrule(lr){8-8}
             & precision                                & recall                                 & f1-score & precision & recall & f1-score & support \\
    \cmidrule(lr){2-4}\cmidrule(lr){5-7}\cmidrule(lr){8-8}
    asian    & 0.86                                     & 0.73                                   & 0.79     & 0.84      & 0.64   & 0.73     & 41861   \\
    black    & 0.70                                     & 0.76                                   & 0.73     & 0.61      & 0.75   & 0.67     & 41904   \\
    hispanic & 0.83                                     & 0.87                                   & 0.85     & 0.80      & 0.84   & 0.82     & 41940   \\
    white    & 0.67                                     & 0.68                                   & 0.68     & 0.57      & 0.53   & 0.55     & 41707   \\
    total    & 0.77                                     & 0.76                                   & 0.76     & 0.70      & 0.69   & 0.69     & 167412  \\
    \bottomrule
  \end{tabular}
  \caption{Accuracy on the test data for the student model after distillation.}\label{tab:student-model}
\end{table}

\subsection{Performance}\label{sec:comparison-performance}

\begin{figure}[ht]
  \centering
  \resizebox{.7\linewidth}{!}{
    \includegraphics[width=\linewidth]{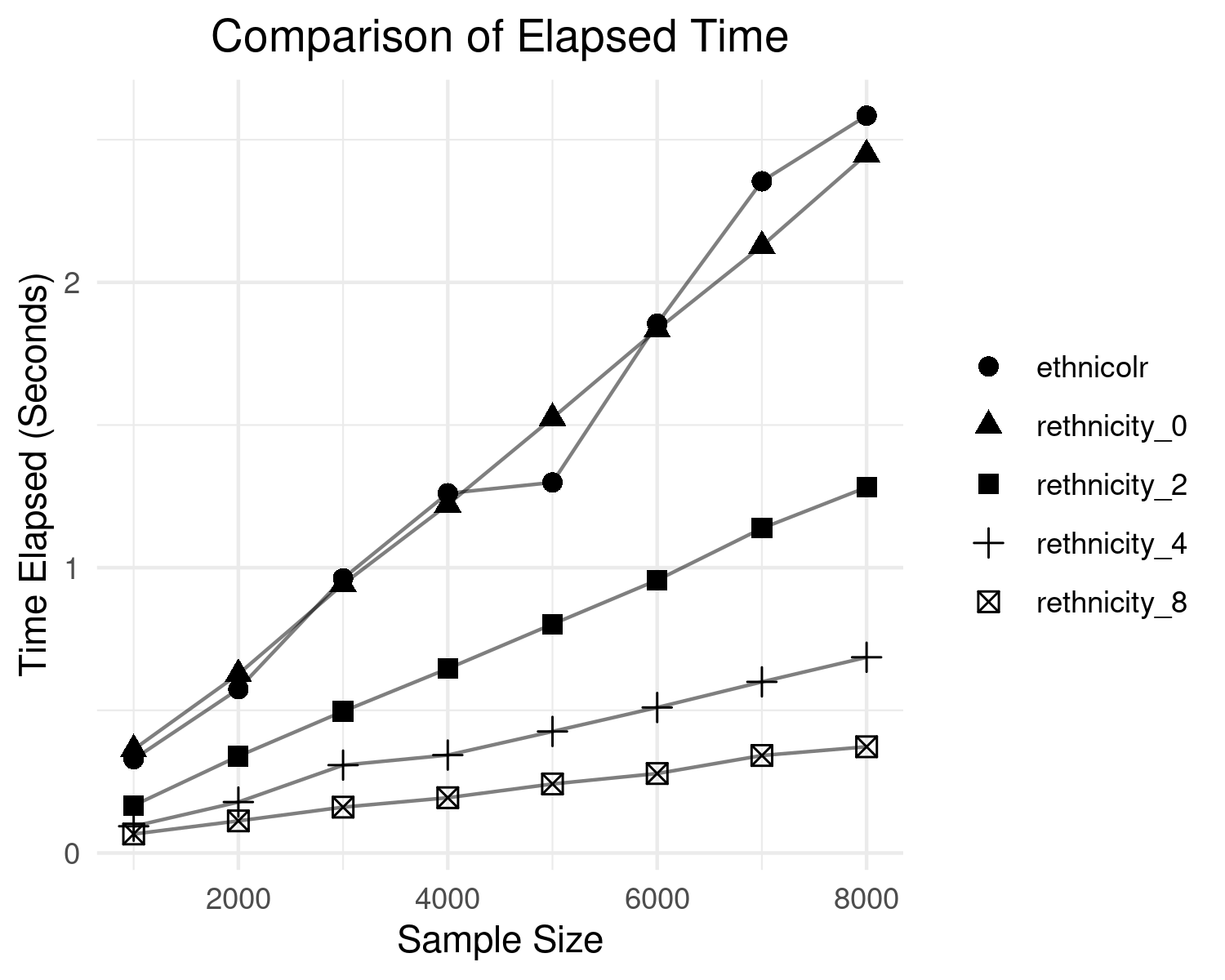}
  }
  \caption{
    Comparison of elapsed time between \texttt{rethnicity} and \texttt{ethnicolr}.
    At any given sample size, the inference is run five times and then the average
    running time is determined as the result for the elapsed time measurement.
    Moreover, a comparison is also made for different numbers of threads available to \texttt{rethnicity}.
    The default single-threaded inference speed is shown as ``rethnicity\_0'' in the plot.
    The performance for inference under two-thread, four-thread, and eight-thread pool is illustrated.
  }\label{fig:comparison-performance}
\end{figure}

The performance of the package is guaranteed by leveraging distillation for model compression,
C++, and multi-threading for low overhead,
as discussed in Sections~\ref{sec:methodology-distillation} and~\ref{sec:feature-performance}.
However, considering speed, there is a need to rigorously test the performance
and compare with it the \texttt{ethnicolr} package as a baseline.

Figure~\ref{fig:comparison-performance} shows that the single-threaded performance
is on par with that of \texttt{ethnicolr}, and the multi-threaded mode achieve further speeds.
First, the distillation method successfully compresses the model and improves performance
by having a smaller model.
The inference speed of the single-threaded distilled model in \texttt{rethnicity} is roughly comparable to
that of the multi-threaded larger model in \texttt{ethnicolr}.
This suggests that the distillation closes the gap between the speedup led by multi-threading
tensorflow\footnote{
  Multi-threading is the default behavior for tensorflow, based on which \texttt{ethnicolr}
  is implemented. \texttt{frugally-deep}, on the other hand, only uses single-thread
  by default.
}.
Second, there is extremely little overhead for multi-threading, and the speedup is almost linear
in terms of the number of threads being used.
The \texttt{rethnicity} package clearly inherits the efficiency of the \texttt{RcppThread} package.
In practice, more threads must be used to process a large dataset,
depending on the size of the dataset and the total number of threads available in the machine.

\section{Using the Package}\label{sec:application}

\subsection{Code Snippet}

The usage of the package is straightforward, as there is only one function provided\footnote{
  More examples can also be found at the GitHub repository:
  \href{https://github.com/fangzhou-xie/rethnicity}{\tt https://github.com/fangzhou-xie/rethnicity}.
}.

\begin{lstlisting}[caption={Example of the \texttt{predict\_ethnicity} function.},captionpos=b]
> predict_ethnicity(firstnames = "Samuel," lastnames = "Jackson")
  firstname lastname prob_asian prob_black prob_hispanic prob_white  race
1    Samuel  Jackson 0.01741119  0.8898849   0.006667824  0.0860361 black
\end{lstlisting}

There are only five arguments for the function \texttt{predict\_ethnicity}: \texttt{firstnames}, \texttt{lastnames},
\texttt{method}, \texttt{threads}, and \texttt{na.rm}.

The \texttt{firstnames} argument accepts
a vector of strings\footnote{
  Character Vector in R.
}, and is only required when \\ texttt{method = `fullname'}.

\texttt{lastnames} also accepts a Character Vector
and is needed for both \texttt{method = `fullname'} and \texttt{method = `lastname'}.

\texttt{method} can only be either \texttt{`fullname'} or \texttt{`lastname'} to indicate whether working only with
last names or both first and last names.

\texttt{threads} can be chosen to have an integer greater than one to leverage multi-threading support for even faster
Data processing\footnote{
  Theoretically, one can choose a number to equal the number of threads in the machine.
  The more threads used, more the overhead introduced in parallel processing, and
  lesser the performance boost gained.
}.

Finally, there is a \texttt{na.rm} argument. This allows one to remove missing values from the input names\footnote{
  For the last name model, only non-missing names are retained for processing and are returned.
  For the full name model, because it requires both first and last names, only names with both
  will be processed.
}.
Otherwise, an error is thrown if values are missing in the input data.
This guarantees that the model has the correct input data and returns meaningful predictions.

\subsection{DIME data}

The DIME dataset offers rich information on the finance and ideology of political campaigns~\citep{bonica2014,bonica2019}.
Following the practice of~\cite{sood2018}, this study also considered this dataset to illustrate one potential usage
of the \texttt{rethnicity} package.

All the donors in the dataset are considered, and their races are predicted using the full-name model, then
the total amount of donation separated by the predicted race is aggregated, and finally, the ratio of donations across
ethnicity is calculated. The results for 2000 and 2010 are  listed in Table~\ref{tab:comparison-dime}.

\begin{table}[ht]
  \centering
  \begin{tabular}{rrrrr}
    \toprule
             & \multicolumn{2}{c}{\bfseries rethnicity} & \multicolumn{2}{c}{\bfseries ethnicolr}                     \\
    \cmidrule(lr){2-3}\cmidrule(lr){4-5}
             & 2000                                     & 2010                                    & 2000    & 2010    \\
    \cmidrule(lr){2-2}\cmidrule(lr){3-3}\cmidrule(lr){4-4}\cmidrule(lr){5-5}
    asian    & 6.29\%                                   & 5.90\%                                  & 2.00\%  & 2.28\%  \\
    black    & 20.83\%                                  & 18.00\%                                 & 8.93\%  & 7.92\%  \\
    hispanic & 4.01\%                                   & 4.44\%                                  & 3.23\%  & 3.31\%  \\
    white    & 68.87\%                                  & 71.66\%                                 & 85.84\% & 86.49\% \\
    \bottomrule
  \end{tabular}
  \caption{
    Comparison of total donations grouped by predicted race from donors' names. The right half of the table
    is taken from \protect\cite{sood2018}.
  }\label{tab:comparison-dime}
\end{table}

Table\ref{tab:comparison-dime} shows that, \texttt{rethnicity} suggests higher ratios of political donation when
compared with \texttt{ethnicolr} results.
This agrees with the accuracies in Section~\ref{sec:comparison-accuracy}
and the discussion on the imbalanced classification problem discussed in Section~\ref{sec:methodology-undersampling},
where \texttt{rethnicity} reduces the error for minority groups significantly.
Without the adjustment, the prediction of white people will be disproportionately higher than that of minority groups,
which underestimates the monetary contribution of minority groups for the elections.

\section{Conclusion}\label{sec:conclusion}

This study demonstrates the methodology and potential usage of the \texttt{rethnicity} package in R.

It leverages different techniques to predict ethnicities. First, undersampling was used to adjust the
imbalance in the racial distribution in the dataset. Second, a character dictionary was used to reduce the
dictionary size and make it independent of training data.
Third, BiLSTM was chosen as the architecture owing to its superior performance in capturing context.
Fourth, after training the gigantic teacher model, the information was distilled by letting it
instruct a much smaller student model.
Finally, the student model was exported to C++ and then loaded via Rcpp.

The model was trained using the Florida Voter Registration dataset using the voters' names, along
with their identified ethnicity. After training the large model,
a smaller student model was also trained and tested.

The objective of building this package was to make the installation and usage easier for
any user interested in predicting ethnicity from names for their research.
The package is entirely native in R with only dependencies being several mature packages
published on CRAN\@. Additionally, it achieves a high performance by delegating heavy computation
to C++ with multi-threading.
The aforementioned advantages are leveraged in the \texttt{rethnicity} package, which is free, fast, and available
to the R community. It also achieves a good performance, particularly for ethnic minorities.

The code snippet is provided as an example of how to use the package. Application to finance and ideology Data
of political candidates is also illustrated.

\section*{Conflict of Interest}

We wish to confirm that there are no known conflicts of interest associated with this publication and there has been no significant financial support for this work that could have influenced its outcome.

\section*{Acknowledgements}\label{}

\bibliography{rethnicity}

\end{document}